\documentclass[runningheads]{llncs}
\usepackage[T1]{fontenc}
\usepackage{graphicx}
\usepackage{amsmath}
\usepackage{amssymb}
\usepackage{multirow}
\usepackage{makecell}
\usepackage{algorithm}
\usepackage{algorithmic}
\usepackage{xcolor}
\usepackage{hyperref}

\begin{document}
\title{Universal Semantic Enhancement for Weakly Supervised Anomaly Detection on Tabular Data}
\titlerunning{Universal Semantic Enhancement for Weakly Supervised Anomaly Detection}
%
%
\author{
Siyu Zhou\inst{1} \and
Yuqin Xie\inst{2}\thanks{Corresponding author.} \and
Hao Yuan\inst{2} \and
Yingjie Zhou\inst{2}
}

\authorrunning{S. Zhou et al.}

\institute{
Department of Computer Science and Engineering,
The Hong Kong University of Science and Technology, Hong Kong\\
\email{szhoubx@connect.ust.hk}
\and
College of Computer Science,
Sichuan University, China\\
\email{xyqyear@gmail.com, yuanhao0607@126.com, yjzhou@scu.edu.cn}
}
\maketitle              
\begin{abstract}
Weakly supervised anomaly detection on tabular data is an essential task in many label-scarce applications, since tabular data is widely encountered and it is commonly observed in practice that only a limited number of labeled anomalies are available alongside abundant unlabeled samples. Most existing methods try to learn features that capture the inherent distribution patterns in the target domain. However, such features often struggle to provide a complete description of possible abnormal characteristics and are highly susceptible to overfitting the extremely scarce labels. To address this challenge, we propose a weakly supervised anomaly detection method based on universal semantic enhancement for tabular data, which fuses the universal knowledge from a pre-trained tabular foundation model with the domain-specific knowledge learned on the target dataset. The two types of knowledge are naturally complementary: universal semantics describe the common patterns of tabular data across domains, while the domain-specific features capture the inherent distributions of normal and anomalous behaviors in the target dataset. A prototype-conditioned modulation mechanism is then designed to effectively integrate the two types of features, allowing them to reinforce each other for improved anomaly detection. Extensive experiments on public datasets show that the proposed method outperforms competitive methods when only eight labeled anomalies are available in the training procedure. Code is available at: \href{https://github.com/Simon-zsy/USE-AD}{https://github.com/Simon-zsy/USE-AD}.

\end{abstract}
\section{Introduction}

Tabular data is one of the most prevalent data forms in real-world applications, and is widely used in areas such as financial risk management, medical diagnosis, and equipment monitoring~\cite{adbench,pang_survey}.
In these areas, detecting anomalies, \textit{i.e.}, the records that deviate from the majority of the data, makes it possible to discover potential risk events such as fraudulent transactions and intrusion behaviors in time, and to reduce the corresponding losses~\cite{pang_survey}.
Building an effective deep anomaly detection model usually requires a large number of labeled samples to guide the training procedure.
However, anomalous events occur rarely in practice, and accurately annotating them requires extensive expert knowledge of the domain, so that collecting a sufficient number of labeled anomalies is extremely expensive~\cite{pang_survey,devnet}.
As a result, practical scenarios usually provide only a small number of labeled anomalies together with abundant unlabeled samples.
Weakly supervised anomaly detection has therefore attracted considerable attention, since it learns detection models from exactly such data and does not rely on large-scale annotation~\cite{devnet,prenet,feawad}.

To make the best use of the scarce labels, most existing weakly supervised anomaly detection methods for tabular data train a feature encoder together with an anomaly scoring function on the target dataset~\cite{devnet,deepsad,feawad,rosas}.
The features learned in this way reflect the inherent distribution patterns of a single domain, namely the domain priors carried by the few labeled anomalies and the statistical patterns of normal behaviors in historical data.
However, anomalies may appear in highly diverse forms, and the features learned from a single domain can hardly give a complete description of these possible forms; in addition, with extremely few labeled anomalies, the learned model tends to overfit these limited samples.
Recently, pre-trained tabular foundation models~\cite{tabpfn} have shown that universal semantics of tabular data can be learned from large-scale and diverse data, and the resulting sample representations describe the common patterns shared across domains.
Such universal knowledge naturally complements the domain-specific knowledge learned on the target dataset, while how to effectively combine the two types of knowledge for weakly supervised anomaly detection remains largely unexplored.
Motivated by this observation, this paper studies how to enhance weakly supervised anomaly detection with universal semantics on tabular data.

In this paper, we propose a weakly supervised anomaly detection method for tabular data based on universal semantic enhancement.
The core idea is to introduce the cross-domain universal semantics contained in a pre-trained tabular foundation model into the domain-specific detection model, so that the universal knowledge and the domain-specific knowledge complement each other.
The two views are complementary: the universal semantic features provide a cross-domain prior that helps find anomalies hard to discover from single-domain data, while the domain-discriminative features capture the normal and anomalous distributions of the target dataset.
Specifically, the proposed method first uses a pre-trained tabular foundation model to extract the universal semantic features for each input sample.
Then, an autoencoder-based feature extractor is trained on the target dataset to obtain the domain-discriminative features, which describe how a sample deviates from the normal patterns of this domain.
Since the two types of features come from different feature spaces and differ in both distribution and semantic level, directly concatenating them tends to cause mutual interference, while fusion modules with many extra parameters are hard to train reliably with the scarce labels.
A prototype-conditioned modulation mechanism is therefore designed, which builds the fusion signal on class prototypes and adaptively integrates the two types of features for each sample; on top of the integrated features, an anomaly scoring network is trained with a weakly supervised loss.
Experiments on various public datasets show that the proposed method outperforms competitive methods, especially when only a few labeled anomalies are available in the training procedure.

The main contributions of this paper are summarized as follows.
\begin{itemize}
  \item We introduce universal semantic enhancement into weakly supervised anomaly detection on tabular data. By bringing in the cross-domain knowledge from a pre-trained tabular foundation model, the proposed method addresses the limitation that existing methods rely only on the priors of a single domain, and allows the universal and domain-specific knowledge to complement each other.
  \item We design a prototype-conditioned modulation mechanism, which seamlessly integrates the universal semantic features and the domain-discriminative features for each sample, and can be trained stably even when the labeled anomalies are extremely scarce.
  \item We conduct extensive experiments on public datasets, and the results show that the proposed method outperforms competitive methods when only a few labeled anomalies are available. Ablation studies further verify the effectiveness of each component.
\end{itemize}

The rest of this paper is organized as follows. Section~2 reviews the related works. Section~3 presents the proposed method in details. Section~4 reports the experimental results and analyses, and Section~5 concludes this paper.

\section{Related Work}
\subsection{Weakly Supervised Anomaly Detection}
Weakly supervised anomaly detection methods learn detection models from a small set of labeled anomalies alongside abundant unlabeled data. DevNet~\cite{devnet} enforces a statistical deviation loss between labeled anomalies and Gaussian reference priors; Deep SAD~\cite{deepsad} extends Deep SVDD~\cite{deepsvdd} to a semi-supervised hypersphere objective; FEAWAD~\cite{feawad} enriches the scoring network with three complementary signals derived from a deep autoencoder. More recently, PReNet~\cite{prenet} amplifies the effective supervision by learning pairwise relation features over labeled pairs, while RoSAS~\cite{rosas} addresses label scarcity by generating continuous-target pseudo-anomalies via interpolation between labeled anomalies and unlabeled samples. Despite these advances, all these methods learn representations only from the target dataset, performing poorly on distributions outside the target data. In the field of visual anomaly detection, researchers have attempted to address this issue using pre-trained models. PatchCore~\cite{patchcore} and WinCLIP~\cite{winclip}, pre-trained on ImageNet~\cite{imagenet} and CLIP~\cite{clip} backends respectively, have outperformed methods trained only on the target data, showing that cross-domain semantic knowledge can be transferred to anomaly detection. However, how to effectively utilize this cross-domain universal semantic knowledge in anomaly detection on tabular data remains an unexplored direction. Our approach aims to fill this gap by leveraging the universal semantics of a pre-trained model on tabular data to enhance the anomaly detection capabilities of domain-specific models.

\subsection{Pre-trained Knowledge Integration and Feature Fusion}
Combining complementary knowledge from heterogeneous sources is a broad strategy across diverse tasks. In vision-language modelling, CLIP~\cite{clip} demonstrates that visual and linguistic features encode orthogonal aspects of the same input, enabling zero-shot generalization that neither modality achieves alone. In natural language processing, augmenting pre-trained representations such as BERT~\cite{bert} with domain-specific features has become standard for low-resource tasks, where the pre-trained encoder supplies cross-domain priors that compensate for scarce labels~\cite{dapt}. For feature fusion, feature-wise linear modulation (FiLM)~\cite{film} achieves cross-modal conditioning via feature-wise affine transformation; more expressive designs, including gated multimodal unit (GMU)~\cite{gmu}, cross-attention~\cite{vaswani}, and DCN v2~\cite{dcnv2} model higher-order interactions at the cost of substantially more trainable parameters. Prototypical networks~\cite{protonet} offer a complementary perspective: anchoring representation learning to non-trainable class prototype vectors enables effective few-shot learning without extensive parameter optimization. At inference time, zero-parameter ensemble methods such as predictive dynamic fusion (PDF)~\cite{pdf} combine complementary prediction scores by weighting each view by its confidence and cross-modal consistency. These existing parameterized fusion modules, however, often fail under extreme label scarcity because a handful of labeled anomalies provides insufficient signal to reliably optimize additional parameters. Our method bridges this gap by grounding the fusion signal in fixed class prototypes, achieving per-sample adaptive modulation without trainable conditioning vectors.

\section{Methods}
\subsection{Problem Formulation}
Let $\mathcal{X} = \{\mathbf{x}_i\}_{i=1}^{n} \subset \mathbb{R}^d$ be a tabular dataset, where $\mathbf{x}_i \in \mathbb{R}^d$ is the feature vector of the $i$-th instance and $y_i \in \{0, 1\}$ denotes its ground-truth label ($y_i = 1$ for anomaly, $y_i = 0$ for normal). In the weakly supervised setting, only a small subset of anomalous samples $\mathcal{M} = \{(\mathbf{x}_j, y_j) \mid y_j = 1\}$ with $|\mathcal{M}| \ll n$ is labeled. The remaining $n - |\mathcal{M}|$ samples form the unlabeled set $\mathcal{U}$, which is predominantly normal but may contain a small fraction of undetected anomalies. The goal is to learn a function $s: \mathbb{R}^d \rightarrow \mathbb{R}$ that maps each instance to an anomaly score, assigning labeled anomalies $\mathbf{x}_p \in \mathcal{M}$ distinctly higher scores than unlabeled samples $\mathbf{x}_q \in \mathcal{U}$, so that the two are clearly separated in the score space.

\subsection{Framework Overview}
%
%
%
%
%
\begin{figure}[t]
\centering
\includegraphics[width=\textwidth]{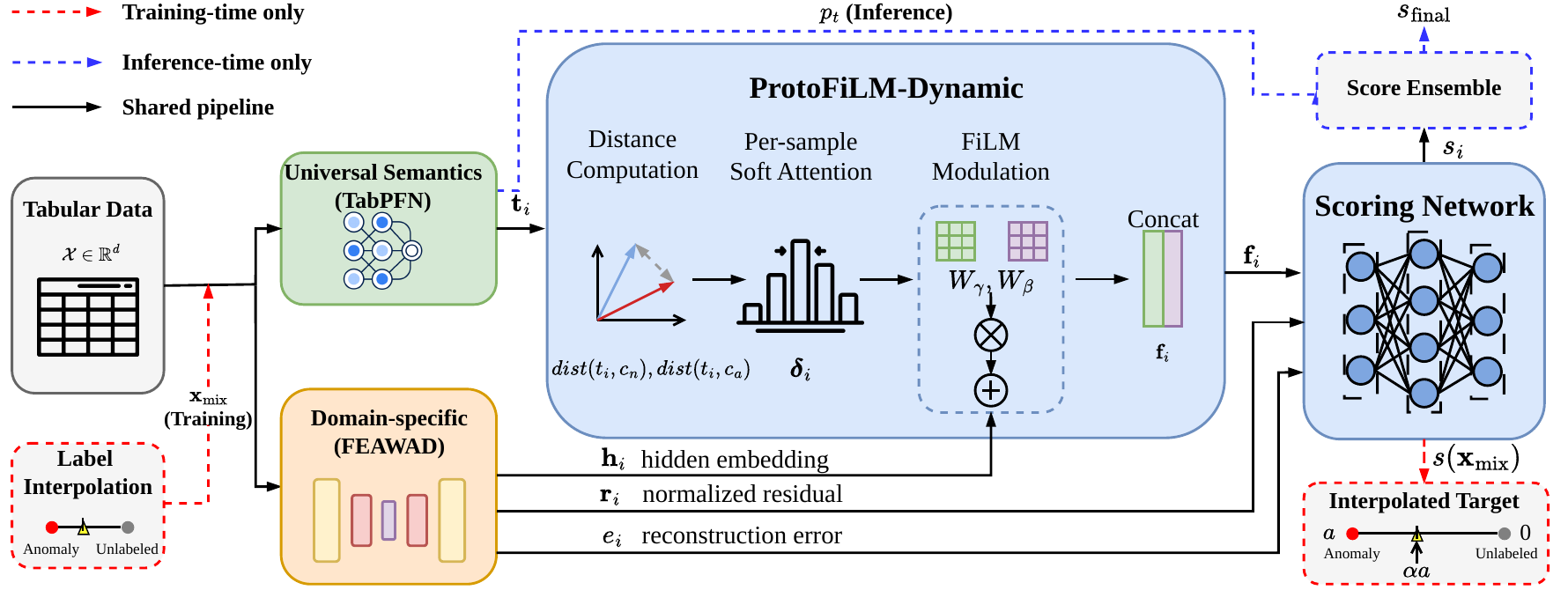}
\caption{Overview of the proposed framework. Universal semantic features $\mathbf{t}_i$ are extracted by the frozen pre-trained model, and domain-specific features $(\mathbf{h}_i, \mathbf{r}_i, e_i)$ by the AutoEncoder. Our ProtoFiLM-Dynamic fuses $\mathbf{h}_i$ and $\mathbf{t}_i$ via per-sample prototype attention to produce the fused representation $\mathbf{f}_i$. The scoring network takes $(\mathbf{r}_i, \mathbf{f}_i, e_i)$ as input and outputs the anomaly score $s_i$. During training, label interpolation between a labeled anomaly and an unlabeled sample with coefficient $\alpha$ generates a pseudo-sample $\mathbf{x}_\text{mix}$ whose score is supervised toward the interpolated soft target $\alpha a$ ($a$ is the score margin); at inference, a zero-parameter score ensemble combines the anomaly score with the pre-trained model's class probability $p_t$ into the final score $s_\text{final}$.}

\label{fig:framework}
\end{figure}

As shown in Figure~\ref{fig:framework}, our framework consists of four components. \textbf{(1) Universal Semantic Feature Extraction}: a frozen pre-trained model on tabular data generates cross-domain embeddings $\mathbf{t}_i \in \mathbb{R}^{512}$ without any gradient updates. \textbf{(2) Domain Discriminative Feature Extraction}: a reconstruction-based feature extractor captures the domain specific knowledge of the target dataset and produces a hidden embedding $\mathbf{h}_i$, a normalized residual $\mathbf{r}_i$, and a reconstruction error $e_i$. \textbf{(3) ProtoFiLM-Dynamic}: our prototype-conditioned modulation module adaptively encodes $\mathbf{h}_i$ via per-sample soft attention over class prototypes, producing the fused representation $\mathbf{f}_i$; during training, a label interpolation strategy generates pseudo-samples $\mathbf{x}_\text{mix}$ with continuous supervision targets between labeled anomalies and unlabeled data, which densifies the scarce supervision signal. \textbf{(4) Anomaly Scoring Network}: a three-layer MLP takes $(\mathbf{r}_i, \mathbf{f}_i, e_i)$ as input and outputs the anomaly score $s_i$; at inference, a zero-parameter score ensemble combines this score with the pre-trained model's class probability $p_t$ according to their confidence and consistency.

\subsection{Module Design}
\noindent\textbf{Universal Semantic Features.}
We choose TabPFN~\cite{tabpfn}, a prior-data-fitting network that is meta-trained on a large number of synthetic tabular classification tasks sampled from Bayesian priors. During testing, it generates calibrated class probabilities by learning context by treating the entire training set as a cue, without any gradient updates. We use its internal representation as a frozen feature extractor: for each sample $\mathbf{x}_i$, we extract
\begin{equation}
  \mathbf{t}_i = \mathrm{TabPFN}(\mathbf{x}_i \mid \mathcal{X}) \in \mathbb{R}^{512},
\end{equation}
where $\mathcal{X}$ denotes the full training set used as in-context examples for TabPFN. Because TabPFN has accessed distributions far exceeding the scope of any single target dataset, $\mathbf{t}_i$ encodes cross-domain statistical regularities, such as which feature combinations are statistically anomalous in multiple domains. More importantly, TabPFN is completely frozen throughout all training phases: its only overhead is a forward propagation of the training set before joint training begins to cache $\mathbf{t}_i$, which reduces our training overhead.

\noindent\textbf{Domain Discriminative Features.}
TabPFN provides universal semantic context; complementarily, we need a feature extractor that exposes how each sample deviates from the structural regularities of the target dataset. We adopt the reconstruction-based feature extraction module from FEAWAD~\cite{feawad}, which is built upon a symmetric encoder--decoder pair: the encoder $f_e: \mathbb{R}^d \rightarrow \mathbb{R}^{64}$ compresses each sample into a compact structural embedding through two fully connected layers ($d \rightarrow 128 \rightarrow 64$), and the decoder $f_d: \mathbb{R}^{64} \rightarrow \mathbb{R}^d$ mirrors it to reconstruct the input through two fully connected layers ($64 \rightarrow 128 \rightarrow d$). Pre-trained on the predominantly normal unlabeled data, the extractor reconstructs normal patterns accurately while producing large reconstruction errors on samples that deviate from them, giving it an inherent capability to expose domain-specific anomalies. From the input $\mathbf{x}_i$ and the reconstructed $\hat{\mathbf{x}}_i = f_d(f_e(\mathbf{x}_i))$, we extract three features:
\begin{align}
  \mathbf{h}_i &= f_e(\mathbf{x}_i) \in \mathbb{R}^{64}, \label{eq:ae_h}\\
  \mathbf{r}_i &= \frac{\hat{\mathbf{x}}_i - \mathbf{x}_i}{\|\hat{\mathbf{x}}_i - \mathbf{x}_i\|_2} \in \mathbb{R}^d, \label{eq:ae_r}\\
  e_i &= \|\hat{\mathbf{x}}_i - \mathbf{x}_i\|_2 \in \mathbb{R}. \label{eq:ae_e}
\end{align}
Each feature plays a unique role: the hidden embedding $\mathbf{h}_i$ carries a compressed structural representation of the sample; the normalized residual $\mathbf{r}_i$ encodes the direction of the reconstruction bias, providing information about the anomaly type; and the scalar error $e_i$ quantifies the magnitude of the bias and is injected as a persistent signal into each layer of the scoring network. Together, these three features provide complementary domain-specific information that may not be fully captured by the semantic representations derived from TabPFN.

\noindent\textbf{ProtoFiLM-Dynamic Fusion}
While universal semantic and domain-specific features are complementary, effectively fusing them under extremely scarce supervision remains a critical challenge. The standard FiLM~\cite{film} applies a global affine transformation to one feature space and conditions it with another feature space, using a single conditional vector shared by all samples. This approach works well in data-rich scenarios but its performance degrades for anomaly detection under severely limited anomaly supervision. 

First, before joint training, we compute two class prototype vectors~\cite{protonet} in TabPFN embedding space. The anomaly prototype $\mathbf{c}_a$ is the mean embedding of labeled anomalies in $\mathcal{M}$; the normal prototype $\mathbf{c}_n$ is the mean of a random subsample $\mathcal{U}_0 \subset \mathcal{U}$, serving as a proxy for the normal class:
\begin{align}
  \mathbf{c}_a &= \frac{1}{|\mathcal{M}|}\sum_{j \in \mathcal{M}} \mathbf{t}_j \in \mathbb{R}^{512}, \label{eq:proto_a}\\
  \mathbf{c}_n &= \frac{1}{|\mathcal{U}_0|}\sum_{j \in \mathcal{U}_0} \mathbf{t}_j \in \mathbb{R}^{512}. \label{eq:proto_n}
\end{align}
Both prototypes are stored as non-trainable buffers throughout joint training.
Second, for each sample we compute squared Euclidean distances to both prototypes in TabPFN space and convert them to soft mixing coefficients via softmax:
\begin{equation}
  [\alpha^n_i,\, \alpha^a_i] = \mathrm{softmax}([-\|\mathbf{t}_i - \mathbf{c}_n\|_2^2,\; -\|\mathbf{t}_i - \mathbf{c}_a\|_2^2]),
  \label{eq:soft_attn}
\end{equation}
where $\alpha^n_i$ and $\alpha^a_i$ are the soft attention weights toward the normal and anomaly prototypes, respectively. We then form the per-sample conditioning vector as their convex combination $\boldsymbol{\delta}_i = \alpha^n_i\, \mathbf{c}_n + \alpha^a_i\, \mathbf{c}_a \in \mathbb{R}^{512}$. Samples close to the anomaly prototype receive $\alpha^a_i \approx 1$; clearly normal samples receive $\alpha^n_i \approx 1$.
Third, two zero-initialized linear layers ($W_\gamma, W_\beta \in \mathbb{R}^{64 \times 512}$) map $\boldsymbol{\delta}_i$ to per-channel scale and shift parameters $\boldsymbol{\gamma}_i = W_\gamma\, \boldsymbol{\delta}_i$ and $\boldsymbol{\beta}_i = W_\beta\, \boldsymbol{\delta}_i$, which modulate the autoencoder hidden embedding:
\begin{equation}
  \mathbf{h}'_i = (1 + \boldsymbol{\gamma}_i) \odot \mathbf{h}_i + \boldsymbol{\beta}_i, \label{eq:hmod}
\end{equation}
where $\odot$ denotes element-wise multiplication. Finally, we get the fused representation $\mathbf{f}_i = [\mathbf{h}'_i;\, \mathbf{t}_i] \in \mathbb{R}^{576}$. Since $\boldsymbol{\delta}_i$ is a convex combination of fixed non-trainable vectors, gradient flow into $W_\gamma$ and $W_\beta$ remains bounded and well-conditioned regardless of label count, which directly avoids gradient starvation.

\noindent\textbf{Anomaly Scoring Network.}
The scoring network receives the normalized residual $\mathbf{r}_i$, the fused representation $\mathbf{f}_i$, and the reconstruction error scalar $e_i$, and generates anomaly scores. The reconstruction error is concatenated in every hidden layer, ensuring that this strong geometric signal remains visible throughout the computation regardless of how the preceding layers transform $\mathbf{r}_i$ and $\mathbf{f}_i$.
\begin{align}
  \mathbf{z}_1 &= \mathrm{ReLU}\bigl(W_1 [\mathbf{r}_i;\, \mathbf{f}_i;\, e_i]\bigr) \in \mathbb{R}^{256}, \label{eq:z1}\\
  \mathbf{z}_2 &= \mathrm{ReLU}\bigl(W_2 [\mathbf{z}_1;\, e_i]\bigr) \in \mathbb{R}^{32}, \label{eq:z2}\\
  s_i &= W_3 [\mathbf{z}_2;\, e_i] \in \mathbb{R}, \label{eq:score}
\end{align}
where all weight matrices are Xavier-initialized~\cite{xavier}. The scalar output $s_i$ is used directly in the deviation loss during training and, at inference, mapped to the open interval $(0,1)$ through the sigmoid function for score fusion.

\subsection{Training and Inference}

Training proceeds in three consecutive phases, as summarized in Algorithm~\ref{alg:training}. In the joint-training phase, each mini-batch $\mathcal{B}$ is class-balanced, containing an equal number of normal samples drawn from the unlabeled set $\mathcal{U}$ and anomalous samples resampled from the labeled anomalies $\mathcal{M}$ (with replacement, since only $|\mathcal{M}|$ are available). The model is trained by minimizing the loss averaged over each batch, with the loss per-sample being:
\begin{equation}
\mathcal{L} = \mathcal{L}_d + \mathcal{L}_e + \lambda\cdot\mathcal{L}_\text{rosas}.
\end{equation}

Part of our objective is inspired by the loss proposed in FEAWAD~\cite{feawad}. For a sample $\mathbf{x}_i$ with label $y_i$, the deviation loss $\mathcal{L}_d$ drives the anomaly score $s_i$ toward $0$ for a normal sample and above the margin $a{=}5$ for a labeled anomaly, and the reconstruction constraint $\mathcal{L}_e$ mirrors this objective on the reconstruction error $e_i$:
\begin{equation}
\mathcal{L}_d = (1-y_i)|s_i| + y_i\max(0, a-s_i),
\end{equation}
\begin{equation}
\mathcal{L}_e = (1-y_i)e_i + y_i\max(0, a-e_i).
\end{equation}

The \textbf{RoSAS interpolation loss}~\cite{rosas} $\mathcal{L}_\text{rosas}$ densifies
the supervision signal with continuous-target pseudo-samples, and is defined on the pairs of anomaly and unlabeled samples formed within the batch $\mathcal{B}$, and each pair is interpolated into exactly one pseudo-sample $\mathbf{x}_\text{mix}$. Specifically, for an anomalous sample $\mathbf{x}_p$ we pair it one-to-one with a distinct normal sample $\mathbf{x}_q$ from the unlabeled half of $\mathcal{B}$, draw an independent coefficient $\alpha \sim \mathrm{Uniform}(0,1)$, and blend the two in feature space into $\mathbf{x}_\text{mix}$:
\begin{equation}
\mathbf{x}_\text{mix} = (1{-}\alpha)\mathbf{x}_q + \alpha\mathbf{x}_p, \label{eq:xmix}
\end{equation}
whose score is regressed toward the soft target $\alpha a$:
\begin{equation}
\mathcal{L}_\text{rosas} = |s(\mathbf{x}_\text{mix}) - \alpha a|, \label{eq:rosas}
\end{equation}
where $s(\mathbf{x}_\text{mix})$ is obtained by passing $\mathbf{x}_\text{mix}$ through the same encoder--fusion--scoring pipeline. Since $\alpha a$ is a continuous soft label rather than a constant, these pseudo-samples populate the whole score interval between the two endpoints $\{0, a\}$ that ($\mathcal{L}_d$ + $\mathcal{L}_e$) supervises, which encourages a smooth scoring function and mitigates overfitting under extreme label scarcity. Each pseudo-sample thus
corresponds one-to-one to a sample of $\mathcal{B}$ and contributes an independent regression term, sharing the network parameters with the batch samples it is built from but without interacting with either of them in the loss.

\begin{algorithm}[t]
\caption{Three-Phase Training Procedure}
\label{alg:training}
\begin{algorithmic}[1]
\REQUIRE Dataset $\mathcal{X}$, labeled anomalies $\mathcal{M}$, hyperparameter $\lambda$
\ENSURE Trained parameters of the whole model
\STATE \textbf{Phase 1: Autoencoder Pretraining} (100 epochs):
\FOR{each batch $\mathcal{B} \subset \mathcal{U}$}
\STATE Update autoencoder by minimizing $\mathcal{L}_\text{ae} = \frac{1}{|\mathcal{B}|}
\sum_{i\in\mathcal{B}}\|\hat{\mathbf{x}}_i - \mathbf{x}_i\|_2^2$
\ENDFOR
\STATE \textbf{Phase 2: Feature Caching}:
\STATE Compute and cache $\{\mathbf{t}_i\} \leftarrow \mathrm{TabPFN}(\mathcal{X})$
\STATE Compute prototypes $\mathbf{c}_a$ from $\mathcal{M}$,\ $\mathbf{c}_n$ from
$\mathcal{U}_0$
\STATE \textbf{Phase 3: Joint Training} (30 epochs):
\FOR{each class-balanced batch $\mathcal{B}$}
\STATE Extract $({\mathbf{h}_i, \mathbf{r}_i, e_i})$ from autoencoder;\ retrieve $\mathbf{t}_i$
from cache
\STATE Compute $\boldsymbol{\delta}_i$ via prototype soft attention (Eqs.~\ref{eq:proto_a}--\ref{eq:soft_attn})
\STATE Modulate $\mathbf{h}_i \rightarrow \mathbf{h}'_i$ via FiLM (Eq.~\ref{eq:hmod});\ form
$\mathbf{f}_i$
\STATE Compute score $s_i$ via scoring network
\STATE For each anomaly \emph{in} $\mathcal{B}$, pair it with a distinct normal sample \emph{in} $\mathcal{B}$;\ draw $\alpha\sim\mathrm{Uniform}(0,1)$, build $\mathbf{x}_\text{mix}$ (Eq.~\ref{eq:xmix}) and score it
\STATE Compute $\mathcal{L} = \mathcal{L}_d + \mathcal{L}_e +
\lambda\cdot\mathcal{L}_\text{rosas}$, averaged over $\mathcal{B}$
\STATE Update autoencoder, $W_\gamma$, $W_\beta$, scoring network via Adam
\ENDFOR
\end{algorithmic}
\end{algorithm}

\noindent\textbf{Inference-Time Score Fusion.}
At inference, two complementary anomaly signals are available: the deviation-based score $s_d = \sigma(s(\mathbf{x})/a) \in [0,1]$, where $\sigma(\cdot)$ is the sigmoid function and $a$ is the score margin, and the TabPFN class probability $p_t \in [0,1]$. We apply predictive dynamic fusion (PDF)~\cite{pdf} as an auxiliary zero-parameter ensemble, which weights each signal by its own prediction confidence and cross-modal consistency:
\begin{align}
w_d &\propto |s_d{-}0.5| + (1{-}|s_d{-}p_t|), \label{eq:wd}\\
w_t &\propto |p_t{-}0.5| + (1{-}|s_d{-}p_t|), \label{eq:wt}\\
s_\text{final} &= \frac{w_d\, s_d + w_t\, p_t}{w_d + w_t}, \label{eq:sfinal}
\end{align}
where $\propto$ denotes proportionality: the two unnormalized weights are defined up to a common factor that cancels when they are normalized by $w_d + w_t$ in $s_\text{final}$, so only their relative magnitude matters. Here $|s - 0.5|$ measures prediction confidence as the distance from the decision boundary, and $(1 - |s_d - p_t|)$ rewards cross-modal consistency between the two views. This fusion introduces no additional training cost and serves as a lightweight complement to the learned framework. Together, the four components and three-phase training procedure form a cohesive framework that systematically addresses cross-domain feature fusion and gradient starvation under extreme label scarcity in weakly supervised anomaly detection on tabular data.

\section{Experiments}
\subsection{Experimental Setup}\label{exp_setup}
We evaluated our model using the ADBench benchmark~\cite{adbench}. ADBench is a standardized dataset containing 57 real-world tabular anomaly detection datasets from various domains, including cyber intrusion, medical diagnosis, and financial fraud. Consistent with previous research, we excluded datasets with fewer than 30 anomalies, resulting in 42 datasets for evaluation. The 42 datasets span five application domains: \textit{medical diagnosis} (annthyroid, breastw, Cardio, Cardiotocography, Hepatitis, Lymphography, mammography, Pima, thyroid, vertebral, WBC, WDBC, WPBC), \textit{network intrusion} (HTTP, InternetAds, SMTP), \textit{image and signal recognition} (ALOI, landsat, letter, mnist, optdigits, PageBlocks, pendigits, satellite, satimage-2, speech, Waveform), \textit{financial fraud and spam} (campaign, SpamBase, Stamps), and \textit{other scientific and industrial data} (cover, fault, glass, Ionosphere, magic.gamma, musk, shuttle, skin, Wilt, vowels, wine, yeast). Dataset sizes range from a few hundred to tens of thousands of samples, with anomaly ratios between 1\% and 35\%. Each dataset was stratified and split into training and test sets at an 80/20 ratio. The unlabeled training set contains 2\% undetected anomalies, reflecting the practical reality that the unlabeled data is not purely normal but is inevitably contaminated with a small fraction of anomalies. In the training set, we fed the model only $|\mathcal{M}|=8$ labeled anomalies, which is a stricter setting than the commonly used $|\mathcal{M}|=30$~\cite{devnet,feawad,prenet}. All experiments were repeated 10 times with different random seeds, and the mean AUROC and mean AUPRC values are reported. All models ran on NVIDIA RTX 3090 GPUs and were trained using the Adam optimizer~\cite{adam} ($\beta_1=0.9$, $\beta_2=0.999$, $\varepsilon=10^{-8}$, $\text{lr}=10^{-3}$). The autoencoder was pre-trained for 100 epochs, and the whole framework is jointly optimized for 30 epochs.

We compare against three categories of baselines. \textbf{(1) Traditional ML classifiers}: Random Forest (RF)~\cite{rf} and CatBoost~\cite{catboost} are trained as supervised binary classifiers, treating the 8 labeled anomalies and all unlabeled training samples as positive and negative classes respectively. \textbf{(2) Deep weakly supervised methods}: DevNet~\cite{devnet} learns an anomaly scoring function via a deviation loss with a Gaussian prior; DeepSAD~\cite{deepsad} extends a hypersphere objective to the semi-supervised setting; FEAWAD~\cite{feawad} enriches the scoring network with three autoencoder-derived feature signals; PReNet~\cite{prenet} amplifies supervision by learning pairwise relation features over labeled pairs. \textbf{(3) Foundation model baseline}: standalone TabPFN~\cite{tabpfn} is applied directly as an in-context classifier.

\subsection{Performance Comparison with Competing Methods}

We evaluate our framework against 7 baselines on all 42 datasets with $|\mathcal{M}|=8$ labeled anomalies, covering three categories as stated in Section \ref{exp_setup}. Tables~\ref{tab:main} and~\ref{tab:main_auprc} report per-dataset AUROC and AUPRC on 20 representative datasets.

\begin{table}[tp]
\centering
\caption{Per-dataset AUROC on 20 representative ADBench datasets ($|\mathcal{M}|=8$, 10 runs). \textbf{Bold}~=~best, \underline{underline}~=~second best. OOM~=~GPU out of memory. Avg.~Rank~(20) is computed over the 20 datasets shown, and Avg.~Rank~(42) over all 42 evaluated datasets.}
\label{tab:main}
\resizebox{\textwidth}{!}{%
\setlength{\tabcolsep}{7pt}
\renewcommand{\arraystretch}{1.1}
\begin{tabular}{lcccccccc}
\hline
Dataset & RF & CatBoost & DevNet & FEAWAD & PReNet & DeepSAD & TabPFN & \textbf{Ours} \\
\hline
ALOI             & 0.5130 & 0.5270 & 0.5106 & 0.5187 & 0.4953 & 0.5388 & \underline{0.6098} & \textbf{0.6466} \\
Campaign         & 0.5323 & \textbf{0.8741} & 0.5868 & 0.5893 & 0.6809 & 0.6329 & 0.8570 & \underline{0.8668} \\
Cardio           & 0.9256 & 0.9722 & 0.8578 & 0.9469 & 0.9333 & 0.9545 & \underline{0.9870} & \textbf{0.9964} \\
Cardiotocography & 0.8212 & 0.9239 & 0.8060 & 0.8171 & 0.8659 & 0.8709 & \underline{0.9340} & \textbf{0.9424} \\
Fault            & 0.6519 & 0.7366 & 0.5863 & 0.6286 & 0.6086 & 0.7574 & \underline{0.7624} & \textbf{0.7826} \\
Hepatitis        & 0.8385 & 0.8154 & 0.7718 & 0.7000 & 0.7846 & 0.6820 & \underline{0.8564} & \textbf{0.8827} \\
InternetAds      & 0.7989 & 0.8278 & 0.7279 & 0.7049 & 0.6982 & 0.8206 & \underline{0.8587} & \textbf{0.8883} \\
Landsat          & 0.6864 & 0.8385 & 0.7293 & 0.7376 & 0.7412 & 0.8142 & \underline{0.8724} & \textbf{0.8843} \\
Letter           & 0.6692 & 0.7487 & 0.5758 & 0.7745 & 0.7426 & 0.9195 & \underline{0.9455} & \textbf{0.9588} \\
Lymphography     & 0.9776 & 0.9724 & 0.9241 & 0.9414 & \underline{0.9897} & 0.9138 & 0.9896 & \textbf{0.9961} \\
Mammography      & 0.6354 & 0.9251 & 0.8058 & 0.8630 & 0.8906 & 0.8591 & \underline{0.9262} & \textbf{0.9386} \\
Pima             & 0.6813 & 0.7715 & 0.6021 & 0.6249 & 0.6126 & 0.6902 & \underline{0.7731} & \textbf{0.8077} \\
Cover            & OOM    & OOM    & 0.8388 & 0.7268 & 0.6806 & 0.9412 & \underline{0.9984} & \textbf{0.9995} \\
SpamBase         & 0.8395 & 0.9185 & 0.8030 & 0.8343 & 0.8949 & 0.7749 & \underline{0.9377} & \textbf{0.9453} \\
Speech           & 0.4951 & 0.6083 & 0.5791 & \underline{0.6174} & 0.5504 & 0.4730 & \textbf{0.6250} & 0.6104 \\
Stamps           & 0.9476 & 0.9602 & 0.9694 & 0.9298 & 0.9723 & 0.9481 & \underline{0.9814} & \textbf{0.9914} \\
Vertebral        & 0.7613 & 0.8258 & 0.7996 & 0.8147 & 0.8504 & 0.7369 & \underline{0.8833} & \textbf{0.9091} \\
Waveform         & 0.6759 & 0.9087 & 0.6915 & 0.8249 & 0.8265 & 0.8187 & \underline{0.9316} & \textbf{0.9475} \\
WPBC             & 0.5038 & 0.5968 & 0.6799 & 0.6319 & 0.7047 & 0.4921 & \underline{0.7242} & \textbf{0.7385} \\
Yeast            & 0.5444 & 0.6073 & 0.5666 & 0.5762 & \underline{0.6261} & 0.6113 & 0.6202 & \textbf{0.6494} \\
\hline
\textit{Avg.~Rank (20)} & 6.4 & 3.7 & 6.5 & 5.6 & 4.9 & 5.3 & \underline{2.1} & \textbf{1.1} \\
\textit{Avg.~Rank (42)} & 6.4 & 3.6 & 6.0 & 5.5 & 5.1 & 5.1 & \underline{2.3} & \textbf{1.1} \\
\hline
\end{tabular}
}
\vspace{0.8em}
\caption{Per-dataset AUPRC on 20 representative ADBench datasets ($|\mathcal{M}|=8$, 10 runs). \textbf{Bold}~=~best, \underline{underline}~=~second best. OOM~=~GPU out of memory. Avg.~Rank~(20) is computed over the 20 datasets shown, and Avg.~Rank~(42) over all 42 evaluated datasets.}
\label{tab:main_auprc}
\resizebox{\textwidth}{!}{%
\setlength{\tabcolsep}{7pt}
\renewcommand{\arraystretch}{1.1}
\begin{tabular}{lcccccccc}
\hline
Dataset & RF & CatBoost & DevNet & FEAWAD & PReNet & DeepSAD & TabPFN & \textbf{Ours} \\
\hline
ALOI             & 0.0350 & 0.0360 & 0.0386 & 0.0354 & 0.0347 & 0.0429 & \textbf{0.0814} & \underline{0.0697} \\
Campaign         & 0.1431 & \textbf{0.4592} & 0.1860 & 0.1886 & 0.2261 & 0.1949 & 0.4173 & \underline{0.4419} \\
Cardio           & 0.8109 & 0.8523 & 0.7716 & 0.8535 & 0.8541 & 0.8069 & \underline{0.9151} & \textbf{0.9204} \\
Cardiotocography & 0.6910 & 0.7978 & 0.6978 & 0.6567 & 0.7333 & 0.7385 & \underline{0.8271} & \textbf{0.8375} \\
Fault            & 0.5106 & \underline{0.6175} & 0.4954 & 0.5305 & 0.5175 & 0.6036 & \textbf{0.6314} & 0.6050 \\
Hepatitis        & 0.6221 & 0.5961 & 0.6239 & 0.4598 & 0.5597 & 0.4396 & \underline{0.7078} & \textbf{0.7143} \\
InternetAds      & 0.6047 & 0.6932 & 0.5681 & 0.5818 & 0.5781 & 0.6476 & \textbf{0.7317} & \underline{0.7053} \\
Landsat          & 0.4440 & 0.6167 & 0.5236 & 0.5517 & 0.4897 & 0.5888 & \textbf{0.6805} & \underline{0.6171} \\
Letter           & 0.1897 & 0.2229 & 0.2377 & 0.3141 & 0.2597 & 0.5925 & \textbf{0.6721} & \underline{0.6253} \\
Lymphography     & 0.8123 & 0.7499 & 0.7302 & 0.8152 & \underline{0.8594} & 0.6420 & 0.8594 & \textbf{0.9605} \\
Mammography      & 0.1848 & \underline{0.5355} & 0.4172 & 0.4485 & 0.4678 & 0.3605 & \textbf{0.5662} & 0.4700 \\
Pima             & 0.5283 & 0.6387 & 0.4873 & 0.5036 & 0.5075 & 0.5541 & \underline{0.6466} & \textbf{0.6624} \\
Cover            & OOM    & OOM    & 0.2744 & 0.0438 & 0.0606 & 0.6449 & \underline{0.8501} & \textbf{0.9643} \\
SpamBase         & 0.8005 & 0.9121 & 0.7648 & 0.7973 & 0.8849 & 0.7112 & \underline{0.9328} & \textbf{0.9335} \\
Speech           & 0.0216 & 0.0581 & 0.0745 & 0.1061 & 0.0926 & 0.0546 & \textbf{0.1099} & \underline{0.1033} \\
Stamps           & 0.8114 & 0.8411 & 0.7894 & 0.7240 & 0.8033 & 0.6286 & \underline{0.8907} & \textbf{0.9363} \\
Vertebral        & 0.4531 & 0.5357 & 0.5475 & 0.5179 & 0.6253 & 0.5013 & \textbf{0.7248} & \underline{0.6962} \\
Waveform         & 0.1021 & 0.3308 & 0.2933 & 0.3470 & 0.3486 & 0.3425 & \underline{0.4688} & \textbf{0.4721} \\
WPBC             & 0.2823 & 0.3841 & 0.4815 & 0.4447 & 0.4906 & 0.2814 & \textbf{0.5408} & \underline{0.5188} \\
Yeast            & 0.3988 & 0.4503 & 0.4505 & 0.4426 & 0.4664 & 0.4475 & \underline{0.4698} & \textbf{0.4702} \\
\hline
\textit{Avg.~Rank (20)} & 6.7 & 4.3 & 5.9 & 5.6 & 4.7 & 5.7 & \underline{1.7} & \textbf{1.6} \\
\textit{Avg.~Rank (42)} & 6.6 & 3.9 & 5.2 & 5.4 & 4.9 & 5.5 & \underline{2.3} & \textbf{1.5} \\
\hline
\end{tabular}
}
\end{table}

Tables~\ref{tab:main} and~\ref{tab:main_auprc} show that our method achieves the best average rank on both AUROC (1.1) and AUPRC (1.6), and this ranking advantage also holds on the full 42-dataset average (1.1 / 1.5). The two most informative reference points are the individual components of our framework: FEAWAD, relying solely on domain-specific features, ranks 5.6 on both metrics, showing that single-domain representations are insufficient under extreme label scarcity; while TabPFN, relying solely on universal semantics, already achieves rank 2.1 / 1.7, substantially better than all domain-only methods. Our method surpasses both, confirming that the two knowledge sources are genuinely complementary. The benefit is most pronounced on datasets where anomalies occupy semantically coherent regions in TabPFN space: InternetAds, Landsat, and Fault yield the largest gains over TabPFN ($+0.0296$, $+0.0119$, $+0.0202$ AUROC respectively), while Cover, Stamps, and Lymphography see our method lead on both metrics simultaneously. The one exception is Speech, where TabPFN outperforms our method on both metrics: Speech anomalies lack stable semantic coherence, so prototype conditioning could introduce a misleading modulation signal under extreme label scarcity, which is a known limitation of prototype-based designs under high intra-class variability~\cite{protonet}.

\subsection{Label Efficiency}

\begin{figure}[t]
\centering
\includegraphics[width=\textwidth]{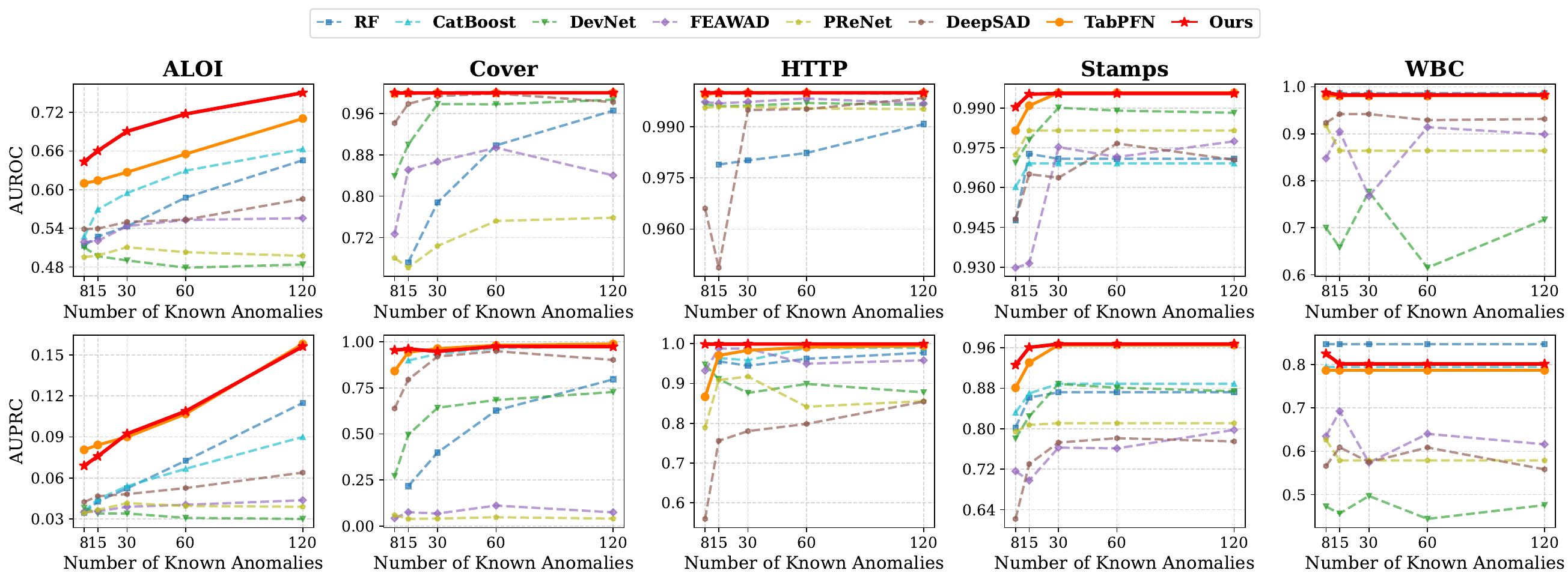}
\caption{AUROC (top) and AUPRC (bottom) vs.\ number of labeled anomalies $|\mathcal{M}|$ on five representative datasets. Ours and TabPFN are shown with solid thick lines; all other baselines are shown with semi-transparent dashed lines.}
\label{fig:label_efficiency}
\end{figure}

To investigate how performance varies with the number of available labeled anomalies, we evaluate all methods with $|\mathcal{M}| \in \{8, 15, 30, 60, 120\}$ on five representative datasets. All other settings follow Section~4.1. Figure~\ref{fig:label_efficiency} reports per-dataset AUROC (top row) and AUPRC (bottom row).

Across the tested values of $|\mathcal{M}|$, our method maintains a clear advantage on most datasets. On Cover and HTTP, our method reaches near-ceiling AUROC and AUPRC even at $|\mathcal{M}|=8$, indicating that universal semantic features supply sufficient cross-domain context to identify anomalies before any domain-specific pattern can be reliably learned from so few labels. On Stamps and ALOI, the gap over FEAWAD is largest at $|\mathcal{M}|=8$ and gradually narrows as more labels are added, which directly confirms that universal semantic enhancement is most valuable precisely when domain-specific supervision is most scarce. The advantage over standalone TabPFN holds across most datasets and supervision levels, indicating that domain-specific autoencoder features provide complementary discriminative signals that remain beneficial even as the number of labeled anomalies increases.

\subsection{Ablation Study}

The goal of this experiment is to quantify the individual contribution of each proposed component and to examine how the components interact. Unlike Section~4.2, which compares our framework against external methods, this section holds all external comparisons fixed and instead ablates each internal design choice one at a time. The ablation baseline is TabPFN-FEAWAD (concat + dev loss, AUROC 0.9174); all other settings are identical to Section~4.2: 42 datasets, $|\mathcal{M}|=8$ labeled anomalies, 10 independent runs. Results are reported in Table~\ref{tab:ablation}.

\begin{table}[t]
\centering
\caption{Component ablation. Each row adds one design choice to the previous configuration. $\Delta$ is relative to the ablation baseline AUROC (0.9174).}
\label{tab:ablation}
\begin{tabular}{lccc}
\hline
Configuration & AUROC & AUPRC & $\Delta$ \\
\hline
Baseline (concat + dev loss)       & 0.9174 & 0.7750 & — \\
+ FiLM (static prototype)          & 0.9178 & 0.7703 & +0.0004 \\
+ ProtoFiLM-Dynamic                & 0.9238 & 0.7775 & +0.0064 \\
+ RoSAS (loss only)                & 0.9183 & 0.7745 & +0.0009 \\
+ PDF (score fusion only)          & 0.9242 & 0.7782 & +0.0068 \\
+ ProtoFiLM-Dynamic + PDF          & 0.9247 & 0.7776 & +0.0073 \\
\textbf{+ ProtoFiLM-Dynamic + RoSAS + PDF} & \textbf{0.9308} & \textbf{0.7799} & \textbf{+0.0134} \\
\hline
\end{tabular}
\end{table}

\begin{figure}[t]
\centering
\includegraphics[width=\textwidth]{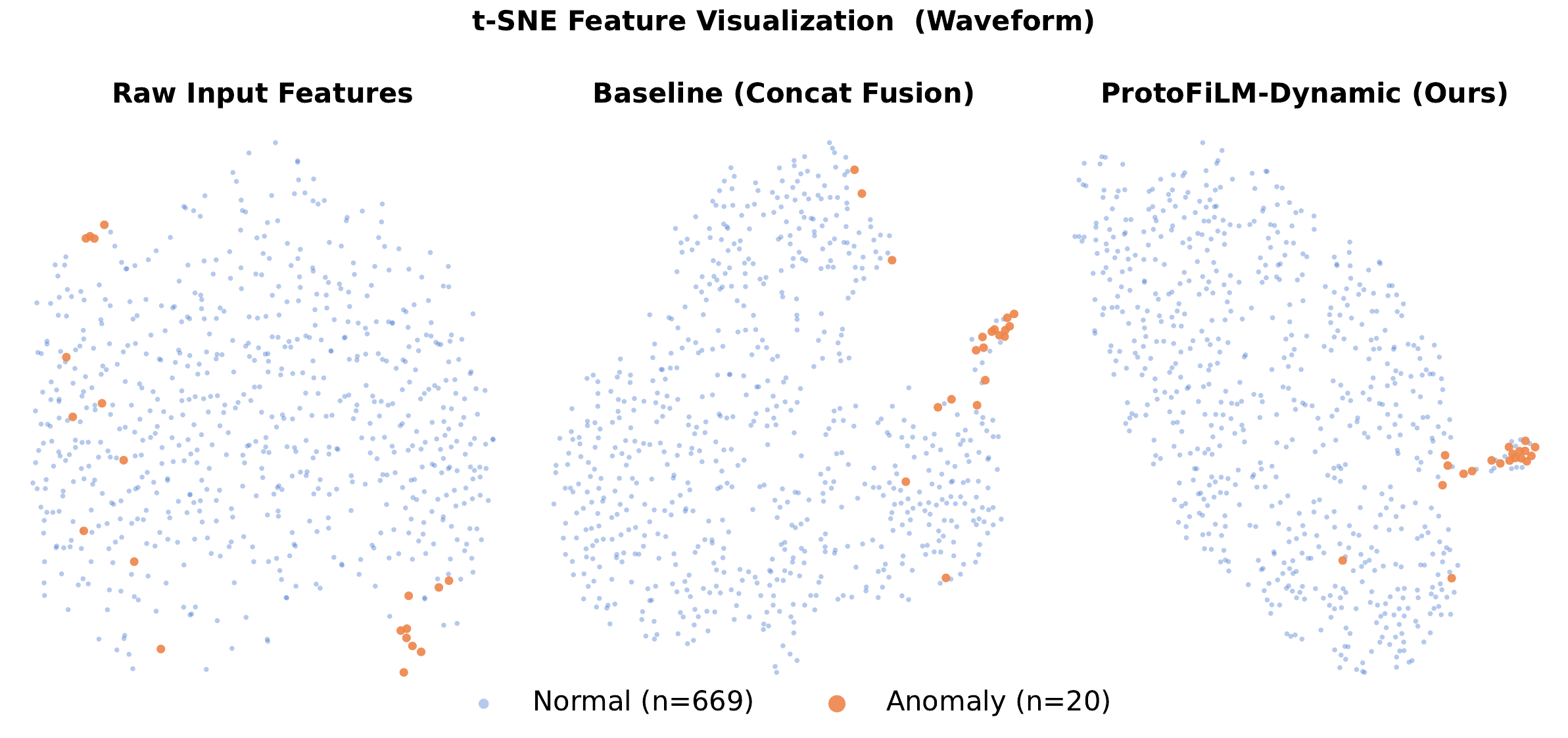}
\caption{t-SNE~\cite{tsne} of the 32-dim penultimate scoring layer on the Waveform dataset ($|\mathcal{M}|=8$, one run). Blue: normal samples ($n=669$); orange: anomalies ($n=20$). Anomaly points in the raw feature space (left) are indistinguishable from normals. The Concat baseline (center) produces partial clustering but with significant normal–anomaly mixing. ProtoFiLM-Dynamic (right) yields a tight, spatially isolated anomaly cluster with a clean decision boundary.}
\label{fig:tsne}
\end{figure}

\noindent\textbf{Fusion Module (ProtoFiLM-Dynamic).}
Replacing ProtoFiLM-Dynamic with a plain concat baseline (row 1 vs.\ row 3) reduces AUROC by 0.0064, confirming that per-sample prototype conditioning is the key driver of the improvement. Figure~\ref{fig:tsne} visualizes the penultimate-layer embeddings and confirms that ProtoFiLM-Dynamic produces a tighter, more separable anomaly cluster than the concat baseline.

\noindent\textbf{Label Augmentation.}
RoSAS applied to the concat baseline alone yields only +0.0009 AUROC, indicating that pseudo-sample interpolation cannot by itself bridge the representation gap. Its contribution becomes substantial in the full system (+0.0134 total), demonstrating that RoSAS is effective as a complement to ProtoFiLM-Dynamic rather than a standalone strategy.

\noindent\textbf{Inference-Time Score Fusion.}
PDF provides an additional gain on top of the full training-time system. This indicates that our model's deviation-based score $s_d$ and TabPFN's class probability $p_t$ are not redundant: even with no extra parameters or training, simply combining the two signals yields a further improvement. The full system (ProtoFiLM-Dynamic + RoSAS + PDF) achieves the best overall performance (+0.0134 AUROC over the ablation baseline).

\subsection{Hyperparameter Sensitivity}

\begin{figure}[t]
\centering
\includegraphics[width=0.6\textwidth]{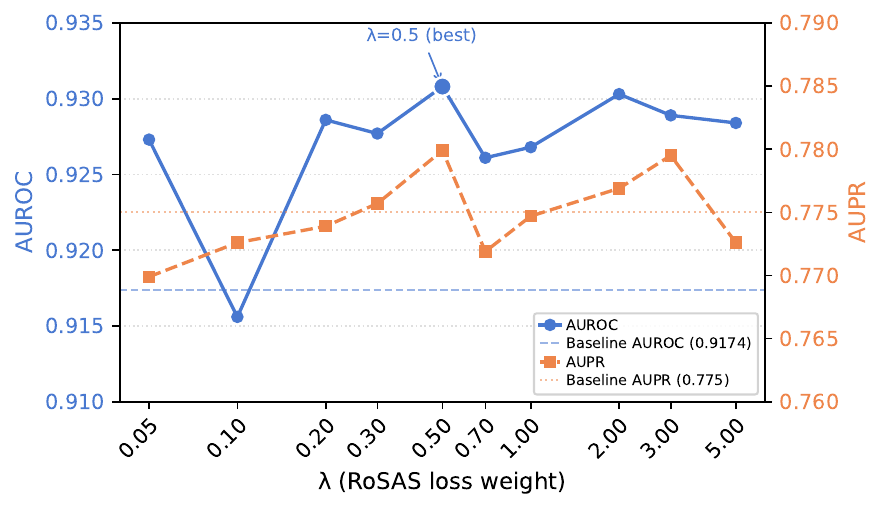}
\caption{Sensitivity of AUROC (left axis, blue) and AUPR (right axis, orange) to the RoSAS loss weight $\lambda$ on 42 datasets ($|\mathcal{M}|=8$, 10 runs). Dashed lines mark the ablation baseline (concat + dev loss) without RoSAS.}
\label{fig:sensitivity}
\end{figure}

The full training objective $\mathcal{L} = \mathcal{L}_d + \mathcal{L}_e + \lambda\cdot\mathcal{L}_\text{rosas}$ introduces a single additional hyperparameter $\lambda$ that controls the relative weight of the RoSAS augmentation loss. Figure~\ref{fig:sensitivity} reports AUROC and AUPR on 42 datasets over 10 runs as $\lambda$ varies across a 100$\times$ range from 0.05 to 5.0.

According to the results, most tested values of $\lambda$ outperform the ablation baseline (concat + dev loss). The curve is non-monotone rather than showing a clear degradation at either extreme: $\lambda\!=\!0.5$ achieves the peak AUROC (0.9308) and $\lambda\!=\!2.0$ is a close second (0.9303), while even the single dip at $\lambda\!=\!0.1$ (0.9156) recovers quickly at $\lambda\!=\!0.2$ (0.9286). This broad optimal region around $0.5$–$2.0$ indicates that the RoSAS signal is beneficial across a wide range of strengths, and the method does not require careful tuning of this hyperparameter. We select $\lambda=0.5$ as the default.

\section{Conclusion}
In this paper, we propose a universal semantic enhancement method for weakly supervised anomaly detection on tabular data, which combines the universal knowledge from a frozen tabular foundation model with the domain-specific features learned by a target-dataset autoencoder. To integrate them under extreme label scarcity, we design ProtoFiLM-Dynamic, a prototype-conditioned modulation module that adaptively enhances each sample by grounding the modulation signal in fixed class prototypes, and adopt a RoSAS-style label-interpolation loss to densify the scarce supervision. On 42 public datasets with as few as eight labeled anomalies, the proposed method attains the best average rank among competitive baselines, with the universal semantics being most beneficial when domain labels are scarcest. Ablation studies further show that the per-sample conditioning of ProtoFiLM-Dynamic is complementary to a parameter-free inference-time score fusion, which brings an additional gain. One limitation of the current design is that class prototypes are computed once before joint training and remain fixed throughout. Future work could focus on online prototype refinement that updates class representations as the scoring network improves, which may further stabilize semantic enhancement under extremely sparse supervision. More broadly, we hope this work motivates further exploration of how universal semantic knowledge from large-scale pretrained models can be systematically integrated with domain-specific representations in other data-scarce detection settings.
%
%
%

\end{document}